\documentclass[pmlr]{jmlr}
\usepackage[utf8]{inputenc}

\usepackage{fnpct}
\usepackage{graphicx}
\usepackage{booktabs}
\usepackage{multirow}
\usepackage{amsmath, amsfonts, bm}
\usepackage{algpseudocode, algorithmicx}
\usepackage{soul} 
\usepackage{lipsum}
\usepackage{mathtools}
\usepackage[shortlabels]{enumitem}

\newcommand{\R}{\mathbb{R}}

\jmlrvolume{230}
\jmlryear{2024}
\jmlrworkshop{Conformal and Probabilistic Prediction with Applications}
\jmlrproceedings{PMLR}{Proceedings of Machine Learning Research}

\title[Calibrated LLMs for Binary Question Answering]{Calibrated Large Language Models for Binary \\ Question Answering}

\author{
    \Name{Patrizio Giovannotti} \Email{patrizio.giovannotti.2019@live.rhul.ac.uk}\\
    \addr{Royal Holloway, University of London, Egham, Surrey, UK}\\
    \addr{Centrica, UK}\\
    \Name{Alex Gammerman} \Email{a.gammerman@rhul.ac.uk}\\
    \addr{Royal Holloway, University of London, Egham, Surrey, UK}
}

\editor{Simone Vantini, Matteo Fontana, Aldo Solari, Henrik Boström and Lars Carlsson}

\begin{document}

\maketitle

\begin{abstract}
Quantifying the uncertainty of predictions made by large language models (LLMs) in binary text classification tasks remains a challenge. Calibration, in the context of LLMs, refers to the alignment between the model's predicted probabilities and the actual correctness of its predictions. A well-calibrated model should produce probabilities that accurately reflect the likelihood of its predictions being correct. We propose a novel approach that utilizes the inductive Venn--Abers predictor (IVAP) to calibrate the probabilities associated with the output tokens corresponding to the binary labels. Our experiments on the BoolQ dataset using the Llama 2 model demonstrate that IVAP consistently outperforms the commonly used temperature scaling method for various label token choices, achieving well-calibrated probabilities while maintaining high predictive quality. Our findings contribute to the understanding of calibration techniques for LLMs and provide a practical solution for obtaining reliable uncertainty estimates in binary question answering tasks, enhancing the interpretability and trustworthiness of LLM predictions.
\end{abstract}


\section{Introduction}
Language models have evolved dramatically, progressing from simple n-gram models to large pre-trained neural networks based on the transformer architecture \citep{vaswani2017attention}. However, their core task remains predicting the next word in a sequence given the previous context. This rudimentary capability has proven remarkably versatile when combined with prompting techniques that allow language models to perform diverse tasks simply by modifying the input text.

For instance, to predict a film review's sentiment using a large language model (LLM), one could construct a prompt:

\begin{quote}
\textit{Read the following review: [...]  The reviewer's opinion is mostly}
\end{quote}

By continuing this prompt, an LLM can generate words like ``positive'' or ``negative'', effectively performing binary sentiment classification without being explicitly trained on that task. This \textit{zero-shot} capability of modern LLMs is powerful, but comes with a critical challenge -- how to reliably quantify the uncertainty of their predictions?

While state-of-the-art LLMs excel at generating fluent and relevant text, their underlying sequence-to-sequence nature makes uncertainty estimation non-trivial. 
This work proposes a simple yet effective approach to extract well-calibrated uncertainty estimates from LLMs for binary question answering tasks, without any further model training or modifications.

The key idea is to directly calibrate the raw word scores (logits) produced by the LLM during text generation. We focus on the logits corresponding to the binary class labels (e.g. ``yes" and ``no") at the first step of generation. By applying Venn--Abers predictors \citep{vovk2022algorithmic-alrw2, Vovk2014VennAbersP} -- a type of conformal predictor providing calibration guarantees under the i.i.d assumption -- we learn an optimal isotonic mapping between these logits and calibrated class probabilities.

We demonstrate the effectiveness of our approach on two binary question answering datasets using the open-source LLM Llama 2 7B \citep{touvron2023llama}. A key advantage is that no further model training -- i.e. any modification to the model's weights as a result of observing examples relevant to the task -- is required, making our method a zero-shot solution for uncertainty-aware binary text classification with LLMs. We also compare against temperature scaling \citep{pmlr-v70-guo17a} and show improved calibration performance.

The remainder of this paper is structured as follows: Section \ref{background} provides background information, Section \ref{methods} describes the proposed methodology in detail, Sections \ref{experimental-setup} and \ref{results} present the experimental setup and results, Section \ref{related-work} comments related work, Section \ref{conclusion} concludes the paper and outlines potential future research directions.

\section{Background}\label{background}
Formally, the language modelling task (see \citealp{jurafsky_speech_2009}) is to compute the probability of a given sequence of words $P(w_{1:n})=P(w_1,w_2,\dots,w_n)$, $w_i\in W\;\forall{i}=1,\dots,n$. This relies on computing the probability of each word $w_i$ given the previous words:
\[
P(w_{1:n}) = \prod_{i=1}^n P(w_i \mid w_{1:i-1})\;.
\]
Estimating directly such a probability is impossible, given the diversity and continuous evolution of human language; however, there are many ways to approximate its value: the simple \textit{bigram} model, for instance, is based on the Markov assumption $P(w_{i}\mid w_{1:i-1})\approx P(w_i\mid w_{i-1})$, with the right-hand side calculated as the proportion of occurrences of the word $w_i$ following word $w_{i-1}$ in a large corpus of text. Current state-of-the-art models for language modelling and text generation, on the other hand, use large \textit{decoder} architectures which are pre-trained on predicting the next word over massive text corpora. Built  upon the attention mechanism \citep{SutskeverSeqToSeq} and often requiring the learning of billions of parameters, decoders were introduced as a component of the first transformer architecture \citep{vaswani2017attention}, but quickly grew to become the foundation of many successful autoregressive generative models, such as the GPT family \citep{Radford2019GPT2}. By effectively estimating the conditional probabilities of words following a given context, generative models are flexible enough to generate coherent text in response to a prompt submitted by the user. 

\paragraph{Tokenization}\label{tokenization}
For text modelling purposes, language have too many words: sampling one out of all possible words of a language at every step is computationally demanding and does not address the presence of unknown words. Instead, LLMs consider sub-words, also known as \textbf{tokens}, that can be part of a word or full words, depending on their frequency in a large reference corpus of text. For example, using SentencePiece, Llama 2's default tokenizer \citep{kudo-richardson-2018-sentencepiece}, the word ``positive'' is represented as a single token \texttt{\_positive}, where the underscore indicates a whitespace preceding the tokenized word. The word ``Positive", on the other hand, is not frequent enough to deserve its own token, so it is represented as two consecutive tokens \texttt{\_Pos} + \texttt{itive}. This strategy helps keep the vocabulary size $K$ manageable, as less frequent words can be represented using combinations of known tokens rather than requiring dedicated ones, but it introduces a layer of complexity whenever we want to use the tokens for other purposes, such as text classification.

\paragraph{Language models as text classifiers}\label{llm-classifiers}
Let $\mathcal{W}=\{w^{(1)}, \dots , w^{(K)}\}$ be a vocabulary of tokens: for example, we could consider the set of all English words post-tokenization. At each text generation step, an LLM outputs a vector $\bm u\in\R^K$, where each component $u_k$ -- called a \textit{logit} -- represents the unnormalized log-probability of token $w^{(k)}$ being the next token in the sequence. These logits can be converted into a probability distribution over the full vocabulary using the softmax function:

\begin{equation}\label{eq:softmax}
P(w^{(k)} \mid \bm{x}) = \frac{\exp(u_k)}{\sum_{j=1}^K \exp(u_j)}
\end{equation}
where $\bm{x}$ is the input sequence. The next token to be generated is then chosen based on a decoding strategy, such as greedy search, beam search, or sampling methods like top-k or nucleus sampling \citep{Holtzman2020nucleus}.

To use an LLM as a classifier, we can provide a list of tokens representing potential labels and prompt the model to select the token that best classifies the input text. However, due to the stochastic nature of text generation, there is no guarantee that the LLM will actually output one of the specified labels, especially for smaller models like Llama 7B.

To address this issue, we propose directly extracting the logits $u_k$ corresponding to the LLM's output tokens at the first step. For example, in a binary question answering task, we extract the logits for tokens representing ``Yes" and ``No", or alternatively their softmax values. We will refer to these tokens as \textit{answer-tokens}. Although these scores are an indication of the token's likelihood to be the true label (or answer), they cannot be directly interpreted as well-calibrated probabilities, since softmax does not guarantee any validity or calibration property.

\paragraph{Calibration}
We refer to \citet{pmlr-v70-guo17a} and define calibration in the following way: given a prediction $\hat Y$ for the label $Y$, returned with an estimated confidence $\hat P$, an ML model is \textit{perfectly calibrated} if
\[
\mathbb{P}(\hat Y = Y\mid \hat P=p)=p, \qquad\forall p\in[0,1]
\]
For instance, let us assume our model made 100 predictions, each with estimated probability $\hat P=0.75$. If the model is perfectly calibrated, exactly 75 out of those 100 predictions need to be correct. In our scenario, a well-calibrated model would output probabilities for the ``Yes'' token that reflect the true rate of positive labels in the test set. Simply applying a softmax function to the raw logits is not enough in most cases, and predictions from LLMs are often poorly calibrated.
Moreover, using softmax does not provide a measure of confidence in the probability estimates themselves -- a property generally enjoyed by imprecise probabilities \citep{pmlr-v179-destercke22a}. 

To overcome these limitations, we employ a recently developed calibration method, which we describe in the following section.

\section{Methodology}\label{methods}
We discuss the two methods we considered to obtain calibrated probabilities as a form of uncertainty estimation: Venn--Abers prediction and temperature scaling. There are many other calibration methods, such as Platt scaling or traditional isotonic regression, however temperature scaling is a popular and widely used technique in the deep learning setting, so we believe it may be the most appropriate competitor for our approach.

\subsection{Venn--Abers Predictors}
The core of our methodology revolves around the use of Venn--Abers predictors for calibration. Venn--Abers predictors \citep{vovk2022algorithmic-alrw2}, a statistical tool used for probabilistic predictions, are employed to adjust the confidence levels of the LLMs’ outputs. We detail the mathematical foundation of these predictors and how they are applied to calibrate the models.

Venn--Abers predictors \citep{Vovk2014VennAbersP} are a special case of Venn predictors, a class of probabilistic predictors guaranteed to be valid under the sole assumption of the training examples being exchangeable. Like all Venn predictors, they 
hold their validity guarantee 
and output multiple probability distributions over the labels -- one for each possible label. The validity property implies perfect calibration (see Figure \ref{fig:reliabubs} for a graphic depiction of a valid model vs a not valid one). 
It has been proven that it is impossible to build a valid probabilistic predictor, in the general sense \citep{gammerman1998}.

As an alternative to the definition given in Section \ref{background}, calibration can be interpreted as follows: let the random variable $Y\in\{0,1\}$ model the label predicted by a binary classifier. Let $P\in[0,1]$ be the confidence associated to the same prediction. $P$ is perfectly calibrated if for the conditional expectation $\mathbb{E}$
\[
\mathbb{E}(Y|P)=P
\]
almost surely.

\begin{figure}[htbp]
\floatconts
    {fig:reliabubs}
    {\caption{Reliability charts for (\emph{a}) Llama 2 7B evaluated zero-shot on our BoolQ test set and (\emph{b}) inductive Venn--Abers predictor based on the same model. The size of the circles represent the proportion of dataset observations falling in a given bin.}}
    {%
        \subfigure{%
            \label{fig:distrib-original}
            \includegraphics[scale=.22]{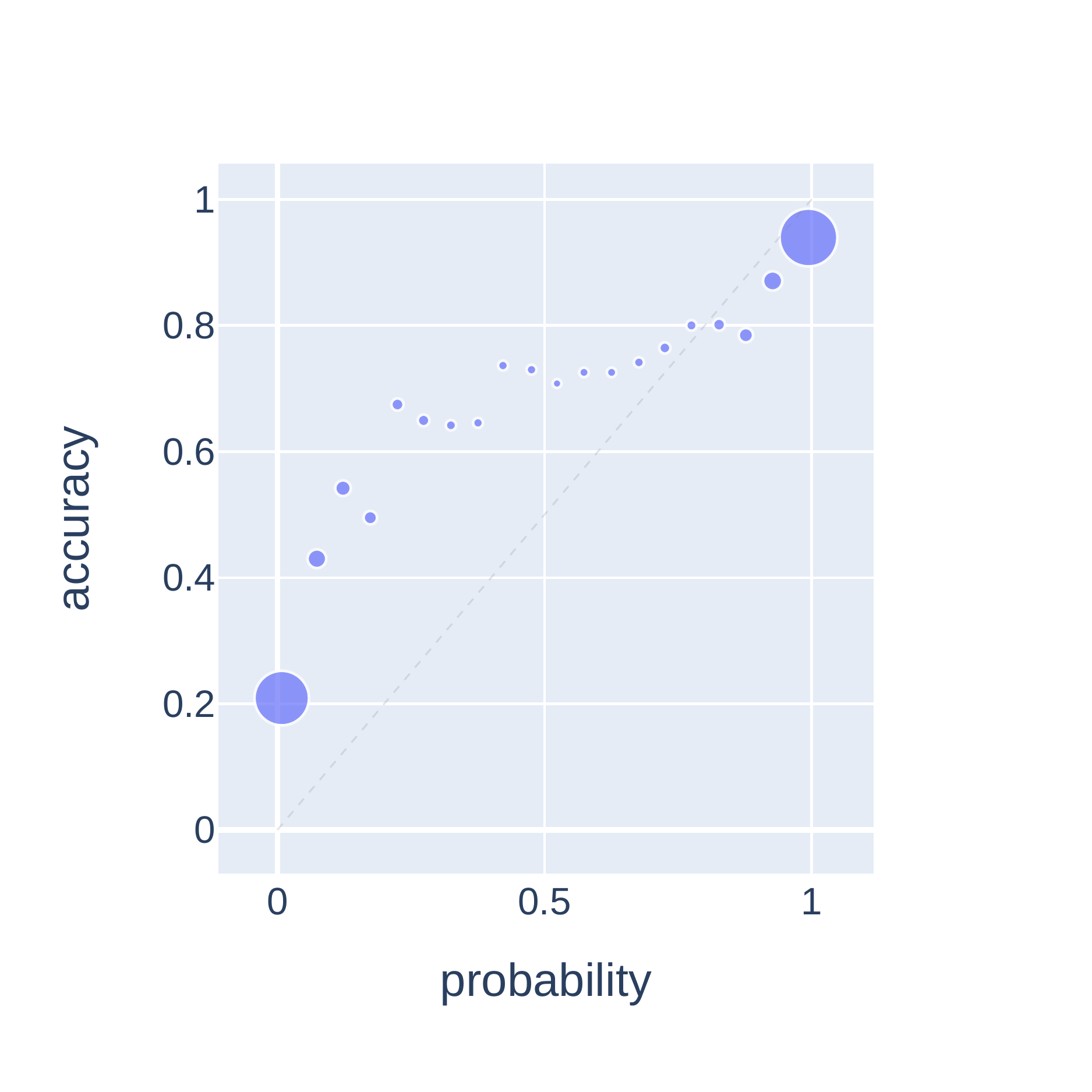}
            }
        \qquad 
        \subfigure{%
            \label{fig:distrib-shuffled}
            \includegraphics[scale=.22]{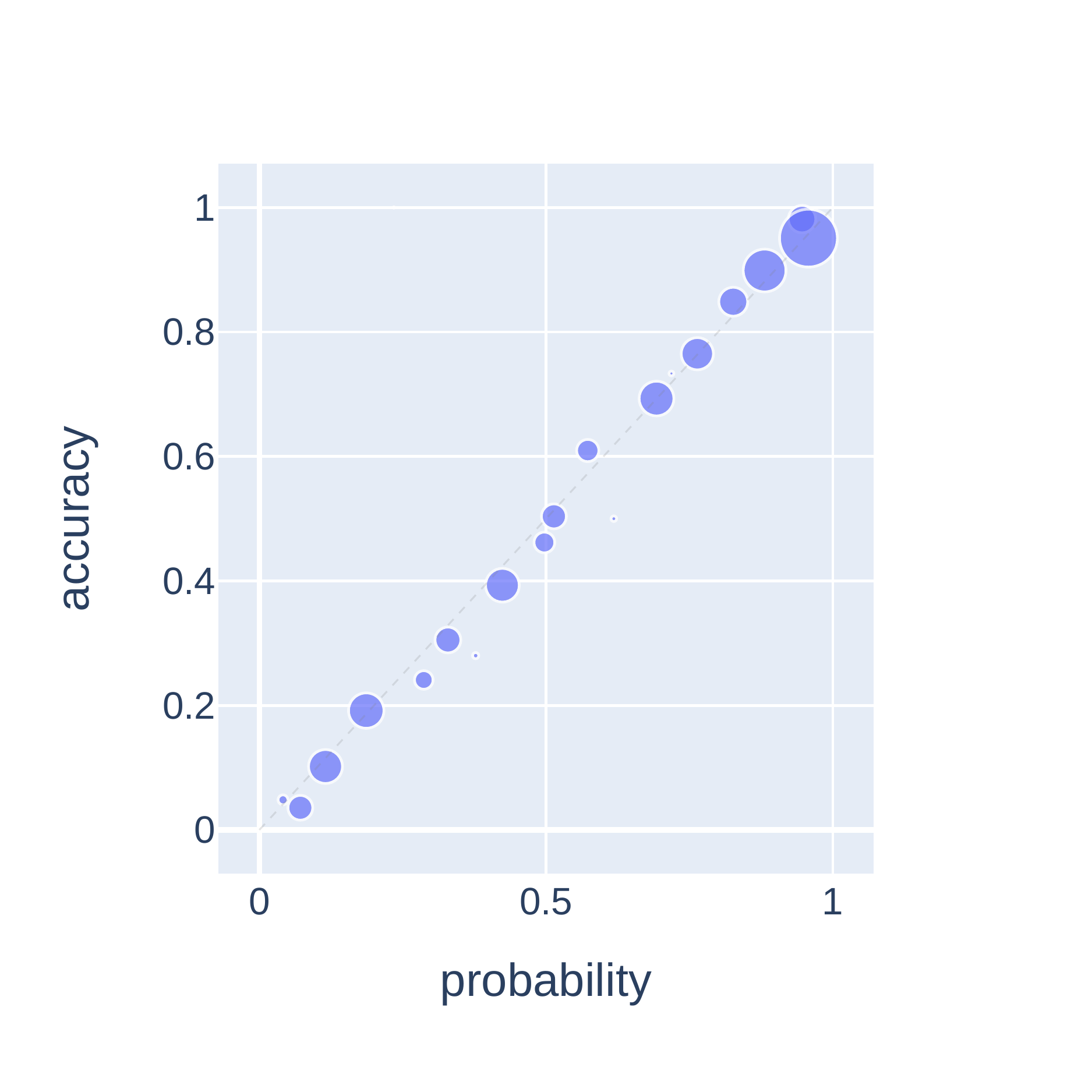}
        }
    }
\end{figure}

Venn--Abers predictors (VAPs) are binary predictors and output a pair of probabilities $(p_0, p_1)$ for each test example $(x,y)$. The former is the probability of $y=1$ should the true label be 0, while the latter is the probability of $y=1$ should the true label be 1: 
one of the two is the valid prediction, but we don't know which one (as we don't know $y$). Because we always have $p_0<p_1$, the pair $(p_0,p_1)$ can be interpreted as the lower and upper probabilities, respectively, of a certain prediction.  Depending on the test example, $p_0$ and $p_1$ may be more or less different in magnitude, although they are usually close to each other. A large gap between $p_0$ and $p_1$ signifies low confidence in the probability estimation -- something traditional probabilistic predictors are not able to provide. For practical reasons however, it is often useful to have one probability estimate per test example. A reasonable way to combine the two numbers, as explained in \citet{Vovk2014VennAbersP}, is to calculate the probability which minimizes the regret for the log loss function:
$$
p=\frac{p_1}{1-p_0+p_1}\:.
$$

In this work we will be using the \textit{inductive} variant of VAPs (IVAP), which was proposed as a computationally lighter version of VAPs in \citet{vovk2015large}. This is our only option as the traditional VAP needs to be retrained for each test example, something absolutely infeasible given the average training time of a transformer model. The only difference with the classical IVAP is that we do not require a proper training set, since the underlying algorithm is pre-trained. This also means we can make use of much more data for calibration and testing.

An IVAP can be created as follows. Suppose we have a binary classification problem and a \textit{scoring algorithm}, i.e. any ML algorithm that can issue a confidence score for each prediction -- in our case, a pretrained transformer $M$. The dataset can be seen as a sequence of $n$ objects $x_i$ labelled as $y_i$, that is, $\mathcal{D}=(x_1, y_1,\ldots, x_n, y_n)$. We divide $\mathcal{D}$ in a calibration set $\mathcal{C}$ of size $m$ and a test set $\mathcal{T}$ of size $n-m$. We run $M$ over all examples in $\mathcal{C}$ and obtain $m$ raw scores (for example,  logits of the answer-tokens). For each test object $x_j$ in $\mathcal{T}$, we predict a score $z_j$ using $M$ and append it to $\mathcal{C}$; then, we fit one isotonic regression on the augmented $\mathcal{C}$ for the case $y_j=0$ and one for $y_j=1$. The resulting probabilities $(p_0, p_1)_j$ are returned for observation $x_j$.

The general procedure to fit an IVAP is given in Algorithm \ref{alg:ivap} (see also \citealp{pmlr-v152-johansson21a}). Isotonic regression is a nonparametric form of regression that fits a step-wise, non-decreasing function to a set of examples (see \citealp{zadrozny2002transforming}).
IVAPs still require for the isotonic regression to be re-calculated for each test example, for each label. However, \citet{vovk2015large} designed an optimised version that requires a single pre-calculation step, then performs an efficient evaluation step for every test example. We use an implementation written in Python \footnote{\url{https://github.com/ptocca/VennABERS}}.
\begin{algorithm2e}[h!]
\caption{Pretrained inductive Venn--Abers predictor}
\label{alg:ivap}
\LinesNumberedHidden
\DontPrintSemicolon
\KwIn{Dataset $\mathcal{D}=(x_1, y_1,\ldots, x_n, y_n)$; pretrained model $M$; calibration size $m$}
\KwOut{Multiprobabilities $((p_0, p_1)_{m+1}, \dots, (p_0, p_1)_n)$}
create calibration set $\mathcal{C}=(x_1,y_1,\dots,x_m, y_m)$ from $\mathcal{D}$\;
create test set $\mathcal{T}=(x_{m+1}, y_{m+1},\dots, x_n,y_n)$ from $\mathcal{D}$\;
\For{$i\leftarrow 1$ \KwTo $m$}{
    compute score for positive label $z_i=M(x_i)$\;
}
\For{$j\leftarrow m+1$ \KwTo $n$}{
    compute score for positive label $z_j=M(x_j)$\;
    fit one isotonic regression $f_0$ on the set $(z_1,y_1),\dots,(z_m,y_m),(z_j,0)$\;
    fit one isotonic regression $f_1$ on the set $(z_1,y_1),\dots,(z_m,y_m),(z_j,1)$\;
    produce the multiprobability $(p_0,p_1)_j=(f_0(z_j), f_1(z_j))$
}
\end{algorithm2e}

\subsection{Temperature scaling}
The softmax function described in Equation \ref{eq:softmax} can be modified with an optional parameter $\tau$, called the \textit{temperature}, which is set in advance and can alter the softmax distribution. Let $\bm u=(u_1,\dots,u_K)$ be the vector of logits returned by the LLM when predicting the next word $w_i$. The probability of word $w^{(k)}$ being chosen at step $i$ is given by the temperature-scaled softmax:

\[
P(w_i=w^{(k)}\mid w_1,\dots,w_{i-1})=\text{softmax}_\tau(u_k)=\frac{\exp(u_i/\tau)}{\sum_{k=1}^K \exp(u_k/\tau)}\;.
\]

Smaller values of $\tau$ (i.e., $\tau < 1$) produce a sharper probability distribution, concentrating most of the probability mass on the most likely words. Conversely, larger values of $\tau$ (i.e., $\tau > 1$) result in a smoother distribution, assigning more probability to less likely words. When $\tau=1$, the temperature-scaled softmax reduces to the standard softmax function.

Temperature scaling \citep{pmlr-v70-guo17a} is a popular calibration method in deep learning. It involves learning a temperature value $\hat{\tau}$ by minimising a calibration loss (e.g., negative log-likelihood) on a separate validation set. The learned parameter $\hat{\tau}$ is expected to approximate the optimal temperature $\tau^*$, which minimises the calibration error on the test set. Temperature scaling is well-suited for deep learning because it employs the same training methodology as the main model and extends naturally to the multiclass setting.

However, temperature scaling has some limitations. Its effectiveness depends on how well the learned temperature $\hat{\tau}$ approximates the optimal temperature $\tau^*$. This approximation relies on two key factors: the similarity between the validation and test distributions, and the effectiveness of the learning algorithm used to estimate $\hat{\tau}$. If the validation set is not representative of the test set, or if the learning algorithm fails to find a good approximation, the calibration performance may degrade. Furthermore, since temperature scaling is a linear transformation of the model's logits, it has an inherent limit on the level of calibration improvement it can achieve, especially if the model's initial calibration is poor.

In contrast, the Venn-Abers predictor always achieves the optimal calibration performance, \textit{irrespective of the temperature}. This property is particularly valuable for LLMs, where users often adjust the temperature to control the generated text's creativity.

\section{Experimental Setup}\label{experimental-setup}
All the experiments are performed using the Llama 2 7B language model, released by Meta as the smallest of the Llama 2 family \citep{touvron2023llama}. Llama 2 7B has a relatively small footprint: it needs about 14 GB of dedicated GPU RAM when making predictions in half precision (16 bit). Because our approach is zero-shot, there is no need for additional 14 GB of memory to store the model gradients for the training step. Most importantly, Llama 2 is an open-source model, and grants access to all its internal components and outputs -- an essential feature of any \textit{white-box} approach (see Section \ref{related-work}). The version used in this work is \texttt{meta-llama/Llama-2-7b-chat-hf}, available on Hugging Face, loaded on a single Nvidia A10G card.

\subsection{Dataset}
Boolean Questions (BoolQ --  \citealp{clark-etal-2019-boolq}) is a question answering dataset for yes/no questions which are produced spontaneously (without specific prompts or directions) by annotators reading a Wikipedia passage. Each example is a triplet $\langle \text{question}, \text{passage}, \text{answer} \rangle$, where the task is to answer a binary question related to the text passage.

In our zero-shot configuration, the original training set is shuffled with the original validation set (the test set is not publicly available), for a total of 12,697 examples. We retain 20\% of it to separately train our Venn--Abers predictor and use the remaining 10,156 examples as test set.

Each example was edited into a prompt that could elicit a satisfactory response from the LLM. Given the relatively small scale of Llama 2 7B, the prompt has been kept as simple as possible. An example of prompt is the following:
\begin{quote}
    \textsf{
    Context:\\
    ``The Air Force usually does not have fighter aircraft escort the presidential aircraft over the United States but it has occurred, for example during the attack on the World Trade Center."\\
    Question: ``Does air force one travel with fighter escort?"\\
    Yes or No?\\
    Answer: }
\end{quote}

\subsection{Evaluation metrics}\label{eval-metrics}
To evaluate calibration performance we use the Expected Calibration Error (ECE).
To compute ECE \citep{naeini2015obtaining}, all predictions are grouped in $M$ bins of equal width, such that bin $B_m$ contains examples with confidence ranging in $(\frac{m-1}{M}, \frac{m}{M}]$.
ECE is defined as 
\[
\text{ECE}\coloneqq \frac{1}{n}\sum_{m=1}^M |B_m|\cdot |p(B_m)-\hat{p}(B_m)|
\]
where $p(B_m)$ is the true fraction of positive instances in bin $B_m$ and $\hat{p}(B_m)$ is the average estimated probability for predictions in bin $B_m$. For example, an ECE of 0.10 means that on average, the models' expected probability for a prediction is off by 10\%. It is important to note that ECE varies depending on the number of bins $M$: throughout our experiments we will report results for $M=10$, which is standard practice in calibration studies -- see for example \cite{pmlr-v70-guo17a}.

To specifically assess prediction quality, we use the area under the ROC curve (AUC), the curve obtained by plotting false positive rate against true positive rate at different classification thresholds. By using AUC, we measure the model's ability of ranking positive examples higher than negative examples, irrespective of the classification threshold and, consequently, irrespective of the model's calibration. Choosing fixed-threshold metrics such as $F_1$ or Matthews Correlation Coefficient would penalise uncalibrated models and hide its actual predictive power.

Additional evaluation metrics are defined and their associated results reported in Appendix \ref{apx:other}.

\section{Results}\label{results}
Following the approach detailed in Section \ref{llm-classifiers}, we extract the logits for both our answer-tokens to predict a binary answer and, consequentially, train our Venn--Abers predictor. We consider two alternative transformations of these scores:
\begin{enumerate}
    \item ``Yes'' and ``No'' scores selected from the softmax over all $K$ logits (\textsf{softmax-$K$})
    \item Scores from softmax computed over the sole ``Yes" and ``No" logits (\textsf{softmax-$2$})
    \item Calibrated version of 1, via the inductive Venn--Abers predictor (\textsf{IVAP-$K$})
    \item Calibrated version of 2, via the inductive Venn--Abers predictor (\textsf{IVAP-$2$})
\end{enumerate}
through a range of temperature values. We consider two pairs of answer-tokens: (\texttt{\_Yes}, \texttt{\_No}) and (\texttt{Yes}, \texttt{No}). The underscore prefix in the first pair indicates that the token is considered a start-of-word token, while the tokens in the second pair can appear in any part of a word (see Section \ref{tokenization}). This subtle distinction is specific to the tokenizer used: a different tokenizer may ignore white spaces and generate the same \texttt{Yes} token regardless of the word's context; in some cases, a token may not even be included in the vocabulary and no logit would be produced as a result. Choosing the right answer-tokens is a delicate early step of our approach and may significantly impact a model's behaviour and performance.

We evaluate our calibration method using expected calibration error and AUC (see Section \ref{eval-metrics}). In Appendix \ref{apx:sst} we report results for a further NLP task, sentiment classification.

\subsection{Calibration results}
In terms of calibration performance, the advantage of using Venn--Abers predictors is evident. Figure \ref{fig:ece} shows ECE values for both answer-token choices. When using start-of-word tokens (\texttt{\_Yes}, \texttt{\_No}), \textsf{Softmax-$K$} shows a minimum at a specific temperature ($\tau\approx1.8$) but degrades rapidly as soon as we move away from it. \textsf{Softmax-$2$}, on the other hand, shows several local minima and a global one for $\tau\approx33$ which outperforms the former model. For the alternative choice (\texttt{Yes}, \texttt{No}), \textsf{Softmax-$2$} shows a global minimum at a relatively low temperature, while \textsf{Softmax-$K$} fails to calibrate the predictions and exhibits high ECE at any temperature. 

In contrast, the Venn--Abers predictors achieve an excellent calibration performance for both token pairs, at any temperature, with the exception of very low values ($\tau<1$) where all models seem to struggle (intuitively, lower temperatures push probabilistic predictions towards the extremes 0 and 1, hence there is little room for the scores to be adjusted).

These findings suggest that while temperature scaling can improve calibration in some cases, it is highly sensitive to the choice of temperature value and may not be effective for all token pairs. On the other hand, the Venn--Abers predictor offers a more reliable and consistent method for obtaining well-calibrated probabilities, making it a promising approach for uncertainty estimation in language models.

\begin{figure}
\floatconts
    {fig:ece}
    {\caption{Expected calibration error of the original Llama 2 model and its Venn--Abers version (IVAP). Our IVAP results in consistently low errors and outperforms temperature scaling, whether we use as labels start-of-string tokens (left) or generic ones (right). IVAP is also invariant w.r.t. how many tokens are considered in the softmax (2 or $K$).}}
    {\includegraphics[width=0.9\textwidth]{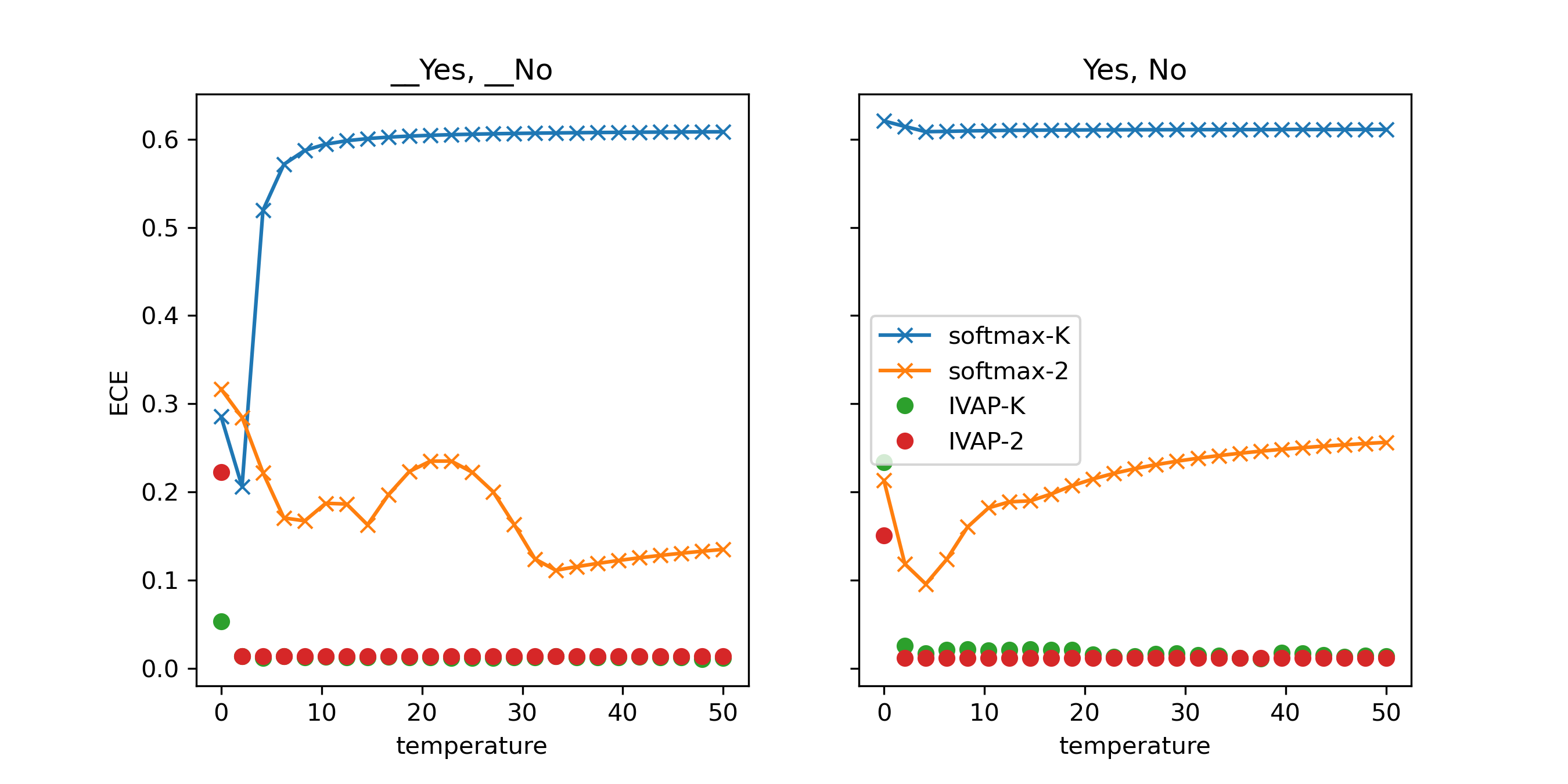}}
\end{figure}


\subsection{Prediction quality}
We report in Figure \ref{fig:AUC} the AUC scores for all models and configurations. We observe immediately that both the original Llama 2 model and the calibrated model obtained via Venn--Abers prediction exhibit similar AUC scores across different temperature settings. This suggests that applying the Venn--Abers predictor does not significantly impact the model's ranking performance, preserving its ability to discriminate between positive and negative examples.

Again, \textsf{Softmax-$2$} (and \textsf{IVAP-$2$}) outperform the two competitors and achieve high AUC for both answer-token choices; \textsf{Softmax-$K$} works better with the (\texttt{Yes}, \texttt{No}) pair, which unfortunately is the configuration where it scored the worse ECE. In contrast, \textsf{IVAP-$2$} was well-calibrated. 

Additionally, we note again that higher temperature values generally result in improved predictive performance, indicating that a more smoothed probability distribution is beneficial for this task.

\begin{figure}
\floatconts
    {fig:AUC}
    {\caption{Area under the ROC curve computed at different temperatures for both models. A positive label was predicted by considering either start-of-word \texttt{\_Yes} tokens (left plot) or generic \texttt{Yes} tokens (right plot).}}
    {\includegraphics[width=0.9\textwidth]{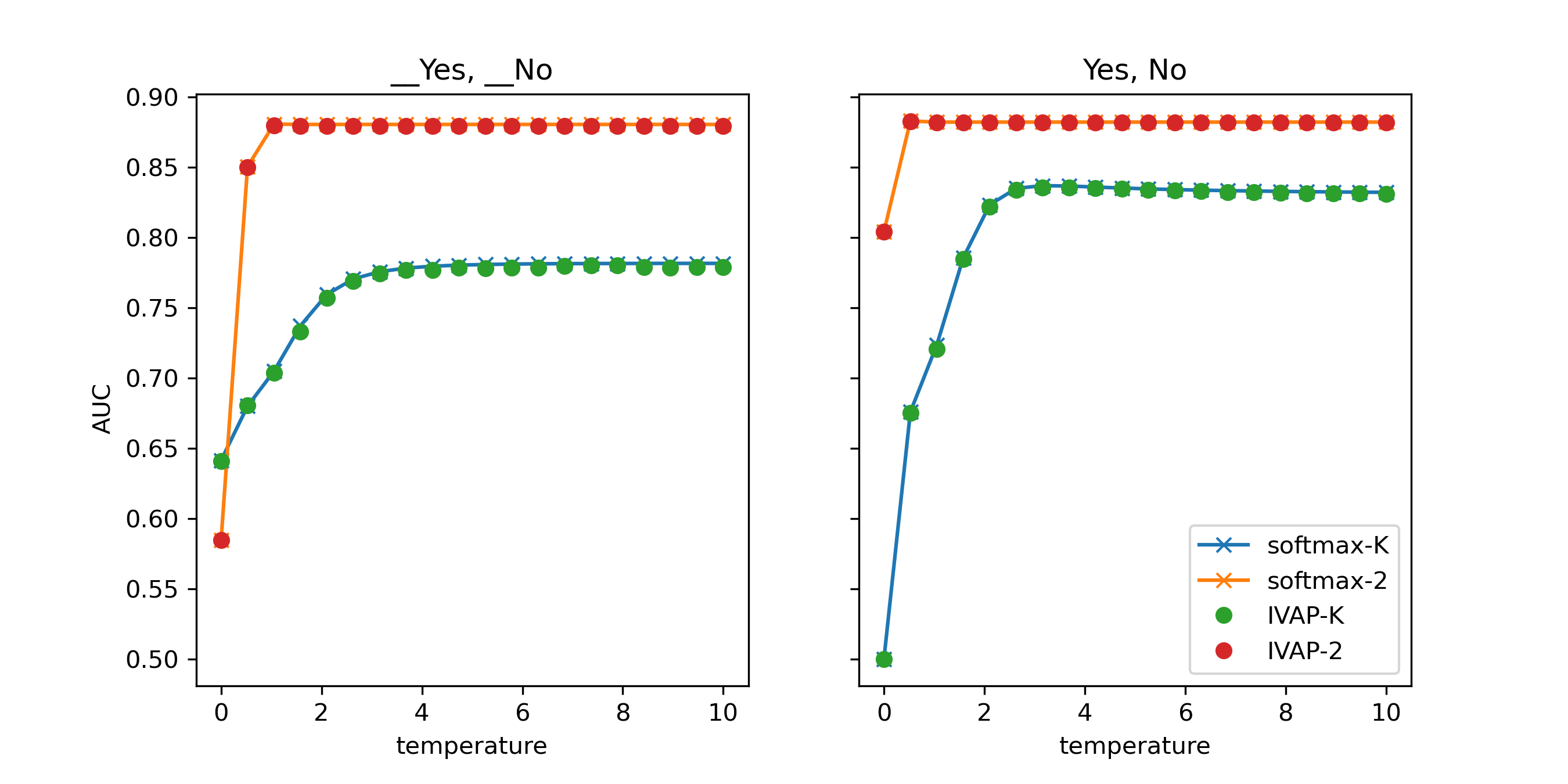}}
\end{figure}

\section{Related Work}\label{related-work}
This work follows the original application of Venn--Abers predictors to pretrained transformers introduced by \citet{pmlr-v179-giovannotti22a}, which we extend to the generative case. Our approach requires access to the internal components of the LLM, namely its output logits, and can be seen as a white-box approach to uncertainty quantification (UQ). \textsc{GPTScore} \citep{fu2023gptscore} is another example of white-box UQ that uses output token weights; other approaches consider the model's internal states \citep{azaria-mitchell-2023-internal} or require a fine-tuning step to learn to express their uncertainty \citep{DBLP:journals/tmlr/LinHE22-teaching}.

Conversely, black-box approaches do not require any knowledge of the model. \citet{kapoor-etal-2024-calibration-tuning} propose a fine-tuning procedure that calibrates the model based on its own evaluation of the generated answer. \citet{manakul-etal-2023-selfcheckgpt}'s \textsc{SelfCheckGPT} computes a confidence score by comparing each LLM claim to $N$ stochastically-generated responses. Together with \citet{Kadavath2022Mostly}'s, this work inspired \citet{agrawal-etal-2024-language} to probe LLMs with different question templates for hallucination detection in the context of reference quotation. \citet{kuhn2023semantic} use an auxiliary model to cluster alternative responses by similarity, \citet{ulmer2024calibrating-only} employs an external model to compute a numerical confidence score, while CRITIC \citep{gou2024critic} can leverage a variety of external tools to validate its output.

Conformal prediction has been recently used in the context of LLM generation: \citet{ravfogel-etal-2023-conformal} showed how to build output token sets containing the correct token at a rate $1-\alpha$; \citet{ulmer-etal-2024-nonexchangeable-knn} extended this \textit{conformal nucleus sampling} strategy to the non-exchangeable case. \citet{su2024api-no-logit} studied the application of CP to black-box models, that is, whenever no access to the logits is available. 
In machine translation, conformal prediction has been used to evaluate translation quality by \citet{pmlr-v204-giovannotti23a} and \citet{zerva2023conformalizing}.

\section{Conclusion}\label{conclusion}
We presented a competitive method to calibrate the output of large language models in the binary question answering setting. Our approach, based on inductive Venn--Abers predictors (IVAP), requires no further training of the LLM and does not require any special assumption on the distribution of the data. 

Our experiments demonstrated that IVAP outperforms a temperature scaling approach and guarantees low calibration error over a broad temperature range. This also applies when choosing different tokens to represent the binary labels. In other words, our approach is invariant with respect to the temperature and to the answer-tokens of choice.

The natural continuation of our work would address question answering with more than two labels, or ideally \textit{open} question answering, where answers can be made of any number of tokens. Additionally, it would be interesting to find the minimum calibration set size that would guarantee an acceptable performance: 1/4 of the test set size may still be too much in certain scenarios.

In conclusion, this is a first step towards a reliable and safer AI, where models can precisely determine and communicate their degree of uncertainty in relation to any answer.

\section*{Acknowledgements}
This work was partially funded by Centrica plc. Thanks to Chris Watkins for clarifying some technical aspects, and to Ilia Nouretdinov for his suggestions and insight.

\appendix

\section{More metrics}\label{apx:other}
For completeness, we evaluated the models using two other metrics for calibration and prediction quality: Brier loss and $F_1$ score (macro-averaged).

The Brier score \citep{Brier-VERIFICATIONOFFORECASTSEXPRESSEDINTERMSOFPROBABILITY} is the mean squared error of the $N$ probabilistic predictions calculated on the test set:
\[
L_B = \frac{1}{N}\sum_{i=1}^N (p_i-y_i)^2
\]
In our case, we have $y_i\in\{0,1\}$ and $p_i$ is the estimated probability of the positive class $P(y_i=1)$. The Brier score loss is preferable to log loss (or cross-entropy loss) for its better handling of high-probability wrong predictions. For example, whenever $p=0$ or $p=1$ is returned for a wrong prediction, log loss would implode to $-\infty$. Results for Brier loss are reported in Figure \ref{fig:brier}, where we notice a similar behaviour to the ECE reported in Figure \ref{fig:ece}. 

\begin{figure}
\floatconts
    {fig:brier}
    {\caption{Brier loss for the original Llama 2 model and the calibrated model using inductive Venn--Abers prediction (IVAP), considering two choices of labeling tokens. }}
    {\includegraphics[width=0.9\textwidth]{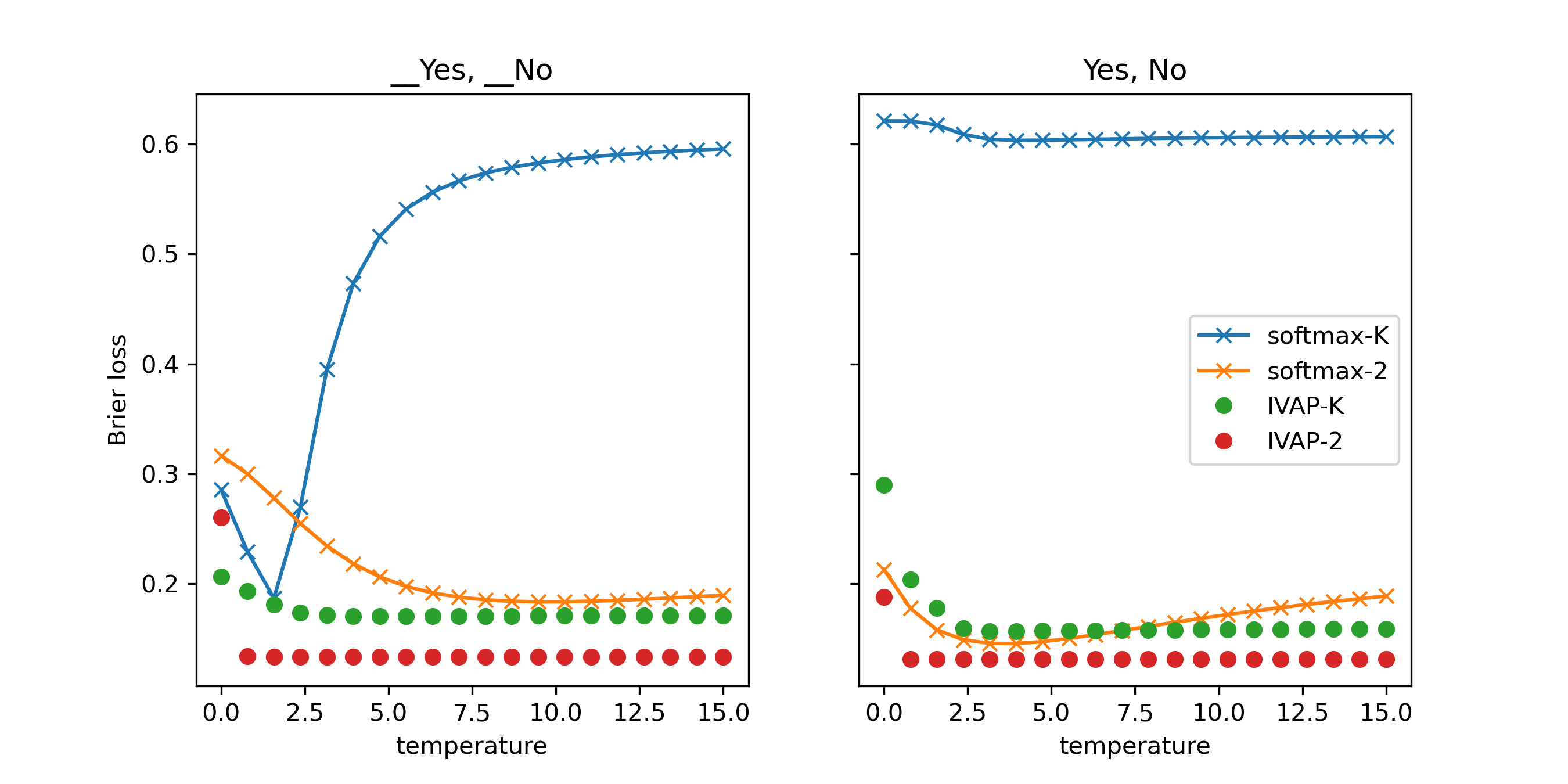}}
\end{figure}

While not the ideal choice for threshold-sensitive scenarios, $F_1$ can simulate an ``out-of-the box'' setting, where the default classification threshold 0.5 is used to give binary answers. Figure \ref{fig:f1} shows that IVAP is still the better choice, almost matched by temperature scaling for a specific choice of tokens and softmax strategy.

\begin{figure}
\floatconts
    {fig:f1}
    {\caption{$F_1$ score for the original Llama 2 model and the calibrated model using inductive Venn--Abers prediction (IVAP), considering two choices of labelling tokens. }}
    {\includegraphics[width=0.9\textwidth]{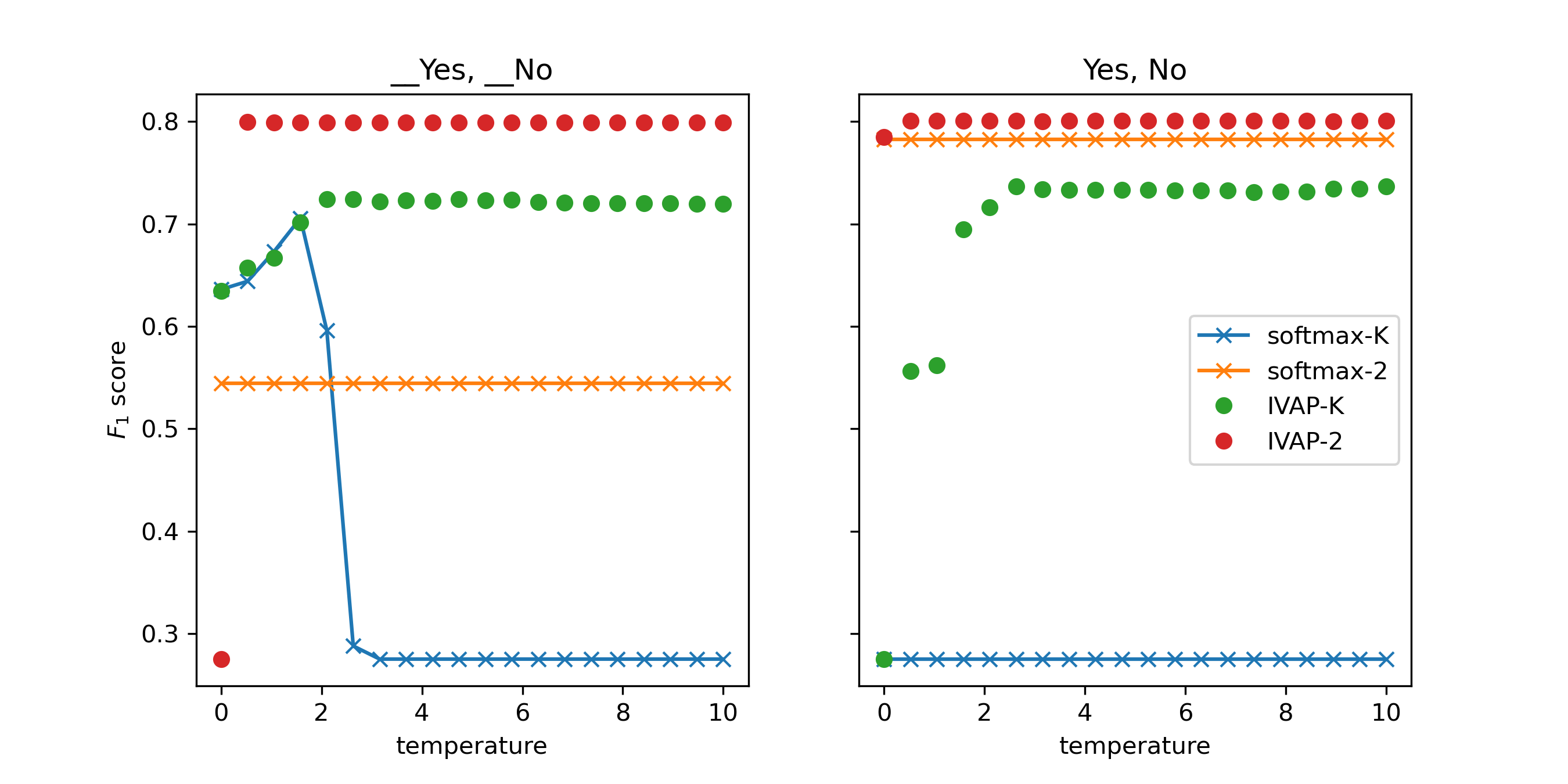}}
\end{figure}

\section{Alternative task: sentiment classification}\label{apx:sst}
We check the effectiveness of our approach against a different NLP task, sentiment classification. For this use case, we use the Stanford Sentiment Treebank \citep{socher-etal-2013-recursive}, a collection of film review excerpts manually labelled with a real number $y\in[0,1]$ representing the reviewer's degree of positive sentiment. We adapt the dataset to the binary case by rounding each label to the nearest integer.

We repeat the same experiments we ran for the BoolQ dataset and find similar results, which we report here. The three default dataset splits were shuffled together and divided again in a calibration set of 2,371 examples and a test set of 9,484 examples. An example prompt is:
\begin{quote}
    \textsf{Film review:\\
    ``Enjoyably dumb, sweet, and intermittently hilarious -- if you've a taste for the quirky, steal a glimpse."\\
    Is the review positive or negative?\\
    Answer:
    }
\end{quote}
We extract the binary answers as described in the paper, using the tokens for ``Pos'' and ``Neg'', which are present in the vocabulary unlike the tokens ``Positive'' and ``Negative''. The results are reported in Figure \ref{fig:ece-sst} and Figure \ref{fig:auroc-sst}, which echo the trends already noticed in the Boolean question answering case, although in this case the token choice actually makes a difference. This is likely due to the fact that there is no \texttt{Neg} token in the vocabulary, so all its scores are set to 0. The \texttt{Pos}, \texttt{\_Pos} and \texttt{\_Neg} tokens are instead available.

\begin{figure}
\floatconts
    {fig:ece-sst}
    {\caption{Sentiment classification task: calibration performance over a range of temperatures. The worse results on the right are likely due to the absence of a speficic \texttt{Neg} token.}}
    {\includegraphics[width=0.9\textwidth]{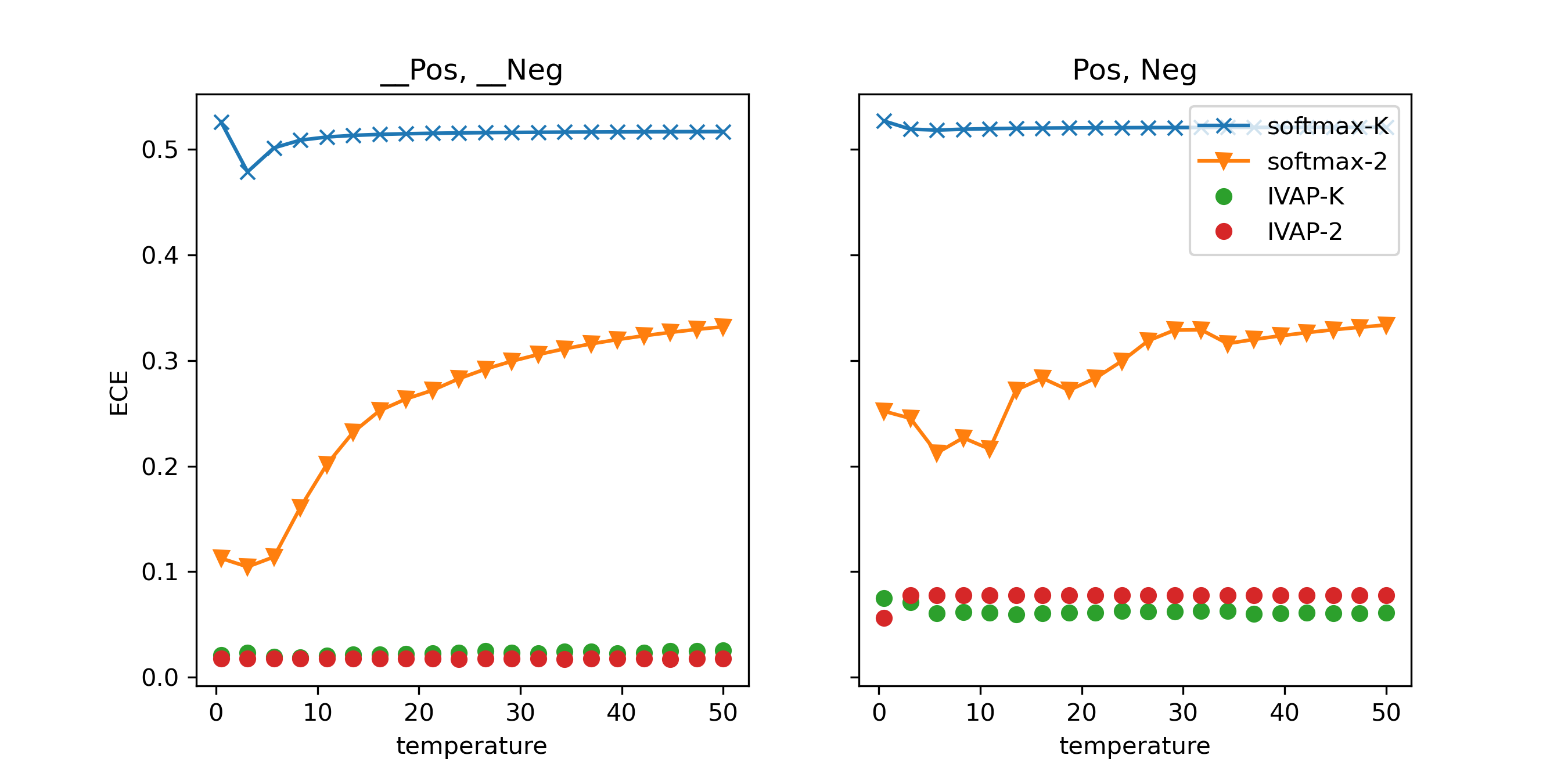}}
\end{figure}

\begin{figure}
\floatconts
    {fig:auroc-sst}
    {\caption{Sentiment classification task: prediction quality (AUC) over a range of temperatures for two labelling token choices (left and right).}}
    {\includegraphics[width=0.9\textwidth]{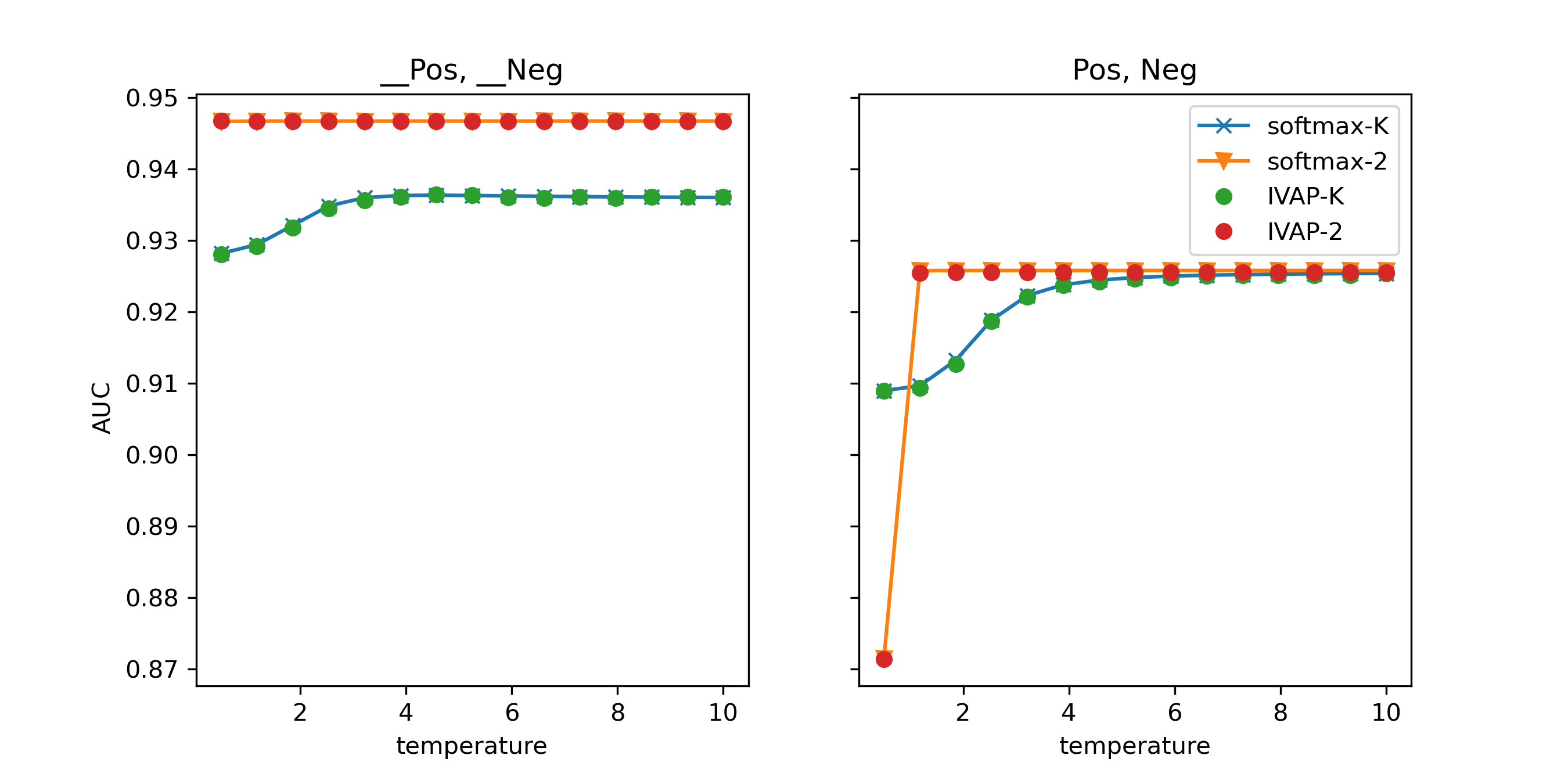}}
\end{figure}

\bibliography{main}

\end{document}